\pdfoutput=1
\documentclass[11pt]{article}

\usepackage{EACL2023}

\usepackage{times}
\usepackage{latexsym}

\usepackage[T1]{fontenc}
\usepackage[utf8]{inputenc}

\usepackage{microtype}

\usepackage{inconsolata}

\usepackage{enumitem}
\setlist[itemize]{noitemsep, nolistsep}
\usepackage{multirow} % check if this package is allowed
\usepackage{booktabs}
\usepackage{csquotes}
\usepackage{amsmath}
\usepackage{tcolorbox}
\usepackage{makecell}

\usepackage{pgfplots}
\usepackage{xurl}
\usepackage{titlesec}

\usepackage{listings}
\usepackage{tablefootnote}
\usepackage{xcolor}
\usepackage{tablefootnote}

\titlespacing{\paragraph}{%
  0pt}{%              left margin
  0.0\baselineskip}{% space before (vertical)
  1em}%   
  
\usepackage{color}
\definecolor{deepblue}{rgb}{0,0,0.5}
\definecolor{deepred}{rgb}{0.6,0,0}
\definecolor{deepgreen}{rgb}{0,0.5,0}
\definecolor{darkgreen}{RGB}{43,163,39}
\definecolor{bluesquare}{rgb}{126,166,224}
\definecolor{LightGray}{gray}{0.9}
\definecolor{DarkGray}{gray}{0.1}

\renewcommand{\tt}[1]{\fontfamily{cmtt}\selectfont #1}

\lstdefinestyle{pythoncode}{
	language=Python,
	otherkeywords={self,join,append,split,write},   % Add keywords here
	keywordstyle=\bfseries\color{deepblue},
	emph={__init__},          % Custom highlighting
	emphstyle=\color{deepred},    % Custom highlighting style
	showstringspaces=false,
	breaklines=true,
	escapeinside=||,
	columns=fullflexible,
	basicstyle=\fontfamily{cmtt}\small,
    belowskip=-\baselineskip,
    aboveskip=-0.7\baselineskip
}

\definecolor{codegreen}{rgb}{0,0.6,0}
\definecolor{codegray}{rgb}{0.5,0.5,0.5}
\definecolor{codepurple}{rgb}{0.58,0,0.82}
\definecolor{backcolour}{rgb}{0.95,0.95,0.92}
\newcommand*\samethanks[1][\value{footnote}]{\footnotemark[#1]}

\title{Causal Reasoning About Entities and Events in Procedural Texts}

\author{Li Zhang$^\clubsuit$\thanks{~~Equal contribution.}, \quad
  Hainiu Xu$^\clubsuit$\samethanks, \quad
  Yue Yang$^\clubsuit$, \quad
  Shuyan Zhou$^\diamondsuit$, \quad \\
  \textbf{Weiqiu You$^\clubsuit$,} \quad
  \textbf{Manni Arora$^\clubsuit$,} \quad
  \textbf{Chris Callison-Burch$^\clubsuit$} \\
  $^\clubsuit$University of Pennsylvania\quad\quad $^\diamondsuit$Carnegie Mellon University \\
  {\tt \{zharry,seacow,yueyang1,weiqiuy,manni,ccb\}@seas.upenn.edu} \\ \tt{\{shuyanzh\}@cs.cmu.edu}
}

\begin{document}
\maketitle
\begin{abstract}
Entities and events are crucial to natural language reasoning and common in procedural texts. Existing work has focused either exclusively on entity state tracking (e.g., \textit{whether a pan is hot}) or on event reasoning (e.g., \textit{whether one would burn themselves by touching the pan}), while these two tasks are often causally related. We propose CREPE, the first benchmark on causal reasoning of event plausibility and entity states. We show that most language models, including GPT-3, perform close to chance at .35 F1, lagging far behind human at .87 F1. We boost model performance to .59 F1 by creatively representing events as programming languages while prompting language models pretrained on code. By injecting the causal relations between entities and events as intermediate reasoning steps in our representation, we further boost the performance to .67 F1. Our findings indicate not only the challenge that CREPE brings for language models, but also the efficacy of code-like prompting combined with chain-of-thought prompting for multihop event reasoning.\footnote{Data and code can be found at \url{https://github.com/zharry29/causal_reasoning_of_entities_and_events}.}
\end{abstract}

\section{Introduction}
Event-centric natural language processing \cite{chen-etal-2021-event} is one of the leading paradigms in machine understanding of texts. This line of work focuses on first extracting entities and events from texts \cite{yang-etal-2019-exploring,du-cardie-2020-event} and then making inferences about them \cite{li-etal-2020-connecting,du-etal-2021-learning}. Even with the recent advances of large language models (LLMs), reasoning about events remains challenging as it requires highly contextual information and ample common-sense knowledge. For example, the event ``\textit{adding water to a pan containing hot oil}'' causes the event ``\textit{there is a sizzling sound}'' to happen, while ``\textit{heat up an empty pan}'' does not. Any model that can draw the correct conclusion given these contexts is expected to have access to some implicit knowledge about these entities and events.

\begin{figure}
    \centering
    \includegraphics[width=\columnwidth]{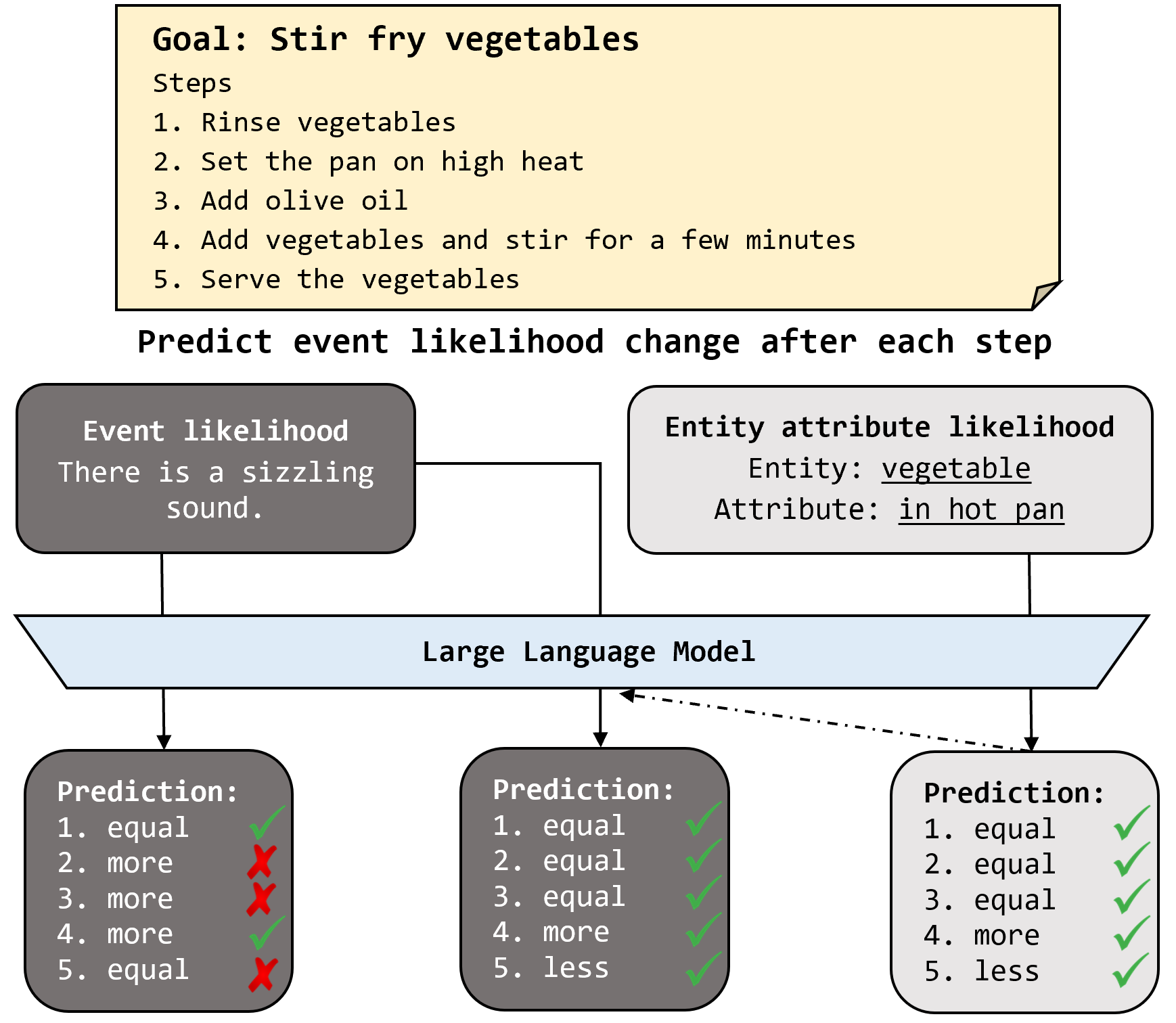}
    \caption{Example of our task CREPE. A procedure including a goal and some steps are provided. A model needs to predict the change in the likelihood of an event throughout the procedure. We show that predicting entity states as an intermediate step improves performance.}
    \label{fig:thumbnail}
    \vspace{-0.3cm}
\end{figure}

One type of text which demonstrates these challenges is procedural text, namely sequences of events, such as how-to instructions, recipes, natural processes, scientific protocols, etc. Procedural texts describe an environment that changes dynamically through a sequence of steps. Therefore, the exact environment configuration is often implicit. In the previous cooking example, whether ``\textit{there is a sizzling sound}'' depends on what steps have taken place. With these interesting challenges coupled with the added benefit of application to robotics \cite{brohan2022can} and household smart assistants such as Alexa \cite{Pennsylvania2022}, reasoning about procedures attracts great attention from the NLP community \cite{https://doi.org/10.48550/arxiv.2205.07455}.

Most work on reasoning about procedural texts has focused solely on either predicting the properties of events (e.g., which event is more likely to happen) \cite{zhang-etal-2020-reasoning, yang-etal-2021-visual, tandon-etal-2019-wiqa} or tracking entity states (e.g., what is some property of an entity after some step) \cite{dalvi-etal-2018-tracking,tandon-etal-2020-dataset}, while the causal relation between events and entities is largely underexplored -- for example, whether ``\textit{there is a sizzling sound}'' is determined by the state of ``\textit{water}'' and ``\textit{oil}.'' Therefore, we claim that many event prediction tasks are multihop reasoning tasks that require the knowledge of intermediate entity states. Causal reasoning about events and entities differs from existing multihop reasoning tasks, such as \citet{yang-etal-2018-hotpotqa,dua-etal-2019-drop} whose reasoning process is explicitly formulated by a direct question (e.g., \textit{how old is the previous US president}); and \citet{geva-etal-2021-aristotle} whose supporting evidence is factual and static. In contrast, causal reasoning in procedures requires models to first figure out the relevant entity attributes, then infer their states based on the current context, and finally predict the event.

To this end, we propose the task of \textbf{C}ausal \textbf{R}easoning of \textbf{E}ntities and \textbf{E}vents in \textbf{P}rocedural Texts (\textbf{CREPE}), with an overview in Figure~\ref{fig:thumbnail}. Given a procedure consisting of a goal~(``\textit{stir fry vegetables}'') and some steps~(``\textit{rinse vegetable}''...), a model is to predict the likelihood of some unobserved events~(``\textit{there is a sizzling sound}'') after the execution of each step. We provide a handcrafted, high-quality benchmark containing 183 procedures, 1219 steps, and 324 changes in the likelihood of events along with the corresponding underlying entity state changes. In an in-context learning setting, we show that most LLMs including GPT-3 \cite{brown2020language} perform no better (.350 F1) than chance (.297 F1), greatly underperforming the human performance of .868 F1, on the development set. Providing ground-truth entity state changes to the prompt of GPT-3 shows no performance gain, indicating that it cannot leverage this causal signal. Instead, we draw inspiration from \citet{madaan22emnlp} who represented texts as programming languages as the prompt to code language model Codex \cite{chen2021evaluating} to perform event reasoning. We propose a novel Python code representation of procedures that achieves .585 F1. Furthermore, our code-like representation allows us to effectively encode and leverage predicted or labeled entity state changes by generating them as an intermediate reasoning step (namely, chain-of-thought), boosting the performance to .667 using predicted entity state changes and .715 F1 using labeled entity state changes.

Our contributions are summarized as follows:
\begin{itemize}[leftmargin=*]
\itemsep0em
    \item We propose a novel task, a dataset, and several strong baselines for causal reasoning about events and entities in procedural texts.
    \item We devise an effective code-like representation of procedures, leading to superior performance and allowing the injection of structured knowledge for reasoning.
    \item We are among the first to show that code language models can apply chain-of-thought to tackle multihop reasoning.
\end{itemize}

\section{Task and Hypothesis}
\label{sec:task}
A procedure $P$ of length $n$ consists of a goal $G$ and some steps $s_1\dots s_n \in S$, each represented as a short sentence. Each procedure is associated with a set of hypothetical events $e_1\dots e_m \in E$ whose likelihood of happening changes throughout the procedure. The task is to predict the change of likelihood of a hypothetical event $e_j$ from step $s_{i-1}$ (the previous step) to step $s_i$ (the current step):
\[\delta_i = p\left(e_j|s_i,\dots,s_1, G\right) - p\left(e_j|s_{i-1},\dots,s_1, G\right)\]
The likelihood change $\delta_i$ is positive if the label is ``more likely'', negative if ``less likely'', or zero if ``equally likely''. 

Predicting the likelihood of hypothetical events, also known as counterfactual reasoning, is extremely important for machine reasoning \cite{10.5555/3238230} (see more in Section~\ref{sec:related_work}). In our work, we hypothesize that \textbf{the causal relation between entity changes and events} can be leveraged by LLMs to better perform counterfactual reasoning. In other words, any change of the likelihood of a hypothetical event is given rise to by changes of some entity attributes $a_1\dots a_m \in A$.
\[\delta_i = p(a_j|s_i,\dots,s_1, G) - p(a_j|s_{i-1},\dots,s_1, G)\]

\section{Dataset}
\label{sec:dataset}
\begin{table}[!t]
    \centering
    \small
    \begin{tabular}{llll}
    \toprule
    \multicolumn{4}{c}{\textbf{Data Statistics}} \\
    \midrule
                                                      & Dev & Test & Total \\ \midrule
    Num. procedures                                   & 42    &  141    &  183     \\
    Num. steps                                        & 295    &  924    & 1219      \\
    Num. event changes & 144    & 180     &  324     \\
    Avg. step per procedure & 7.0 & 6.6  & 6.7 \\ 
    Avg. token per step & 6.8 & 6.8 & 6.8 \\
    \midrule
    \midrule
    \multicolumn{4}{c}{\textbf{Procedure Topics}} \\
    \midrule
                                                      & Dev & Test & Total \\ \midrule
    Recipe                                   & 10    &  33    &  43     \\
    Household                                        & 12    &  40    & 52      \\
    Craft & 4    & 17     &  21     \\
    Technology & 5 & 19  & 24 \\ 
    Travel & 4 & 4 & 8 \\
    Sports & 2 & 13 & 15 \\
    Others & 5 & 15 & 20 \\
    \bottomrule
    \end{tabular}
    \caption{Statistics of the CREPE dataset.}
    \label{tab:dataset_stats}
    \vspace{-.5cm}
\end{table}

Our CREPE benchmark dataset has two portions. The first is handcrafted and cross-validated by six authors of this paper. The annotation happens in 3 phases: (1) we first write down or acquire a procedure from the web; (2) we then annotate some hypothetical events whose likelihood of happening changes throughout the procedure, and how their likelihood change after each step; (3) for each event, we annotate a tuple of entity, attribute, and change that causes the event likelihood change. To obtain interesting and challenging data, we require annotators to write procedures covering a diverse range of topics and to prioritize events that undergo multiple likelihood changes, and those that involve information implicit from the steps. In our work, we strictly use this portion as the development set to inform all our experimental designs.

The second portion, designed to be drawn from a different distribution to minimize bias, was annotated by students in an Artificial Intelligence class at the University of Pennsylvania who participated in an extra-credit assignment. The students were given an overview of the project and some guidelines to annotate data with the aforementioned criteria. We carefully validated all resulting annotations by discarding or editing erroneous and inappropriate examples. In our work, we strictly use this portion as the test set to evaluate the generalization ability of our final models. The complete dataset and annotation instructions can be found in our public repository containing no personally identifiable information of any annotator. 

The statistics of CREPE are in Table~\ref{tab:dataset_stats}. In this work, we consciously focus on few-shot and in-context settings because our data annotation inevitably contains bias and limitation, and thus cannot be truly representative of counterfactual reasoning in every scenario. In such cases, we believe having a sizeable training set aggravates such biases and induces spurious artifacts.

\section{Event Likelihood Prediction}

The task of CREPE is essentially ternary classification, where the likelihood change of each event after each step is labeled as one of “more likely”, “less likely”, or “equally likely”. In this section, all models have no access to the annotated entity state changes until later sections.

\subsection{Baselines}
To show the challenge CREPE brings to existing models, we first introduce some naive baselines.
\begin{itemize}[leftmargin=*]
    \item The \textbf{chance} baseline assigns random labels.
    \item The \textbf{majority} baseline always assigns the majority label “equally likely”.
\end{itemize}

Next, we consider the following state-of-the-art LLMs as strong baselines, where all models are given exactly three examples in their prompt:

\begin{itemize}[leftmargin=*]
    \item \textbf{T5} \cite{2020t5} is one of the state-of-the-art LLMs. Given the goal, steps, and question formatted by a prompt template, we compare the probability of generating ``the answer is no$|$yes.'' We use \texttt{T0-3B}\footnote{\url{https://huggingface.co/t5-3b}} with 3 billion parameters.
    %\medbreak
    \item \textbf{T0} \cite{sanh2022multitask} is a variant of T5, fine-tuned on a large set of downstream tasks with natural language prompts. We adopt the same inference process as T5 described above. We use \texttt{T0pp}\footnote{\url{https://huggingface.co/bigscience/T0pp}} with 11 billion parameters.
    %\medbreak
    \item \textbf{GPT-3} \cite{brown2020language} is a series of LLMs that excels at few-shot learning using the prompting mechanism. We consider \texttt{text-curie-001} (7B parameters), \texttt{text-davinci-002}, \texttt{text-davinci-003}, and \texttt{ChatGPT} (all 175B parameters). We use default parameters with a temperature of 0 for deterministic predictions. An example of the prompt is shown in Figure~\ref{fig:GPT-3_prompt}.
    \item \textbf{GPT-3 finetuned on StrategyQA} is a GPT-3 \texttt{curie} model finetuned with StrategyQA \cite{geva-etal-2021-aristotle}, a dataset of factual multihop questions and their decomposition. StrategyQA is similar to our task in that estimating the change of event likelihood can also be decomposed into sub-tasks of estimating the change of state of related entities (Section~\ref{sec:cot}). 
\end{itemize}

\begin{figure}[t!]
    \centering
    \begin{center}
    \begin{tcolorbox} [top=2pt,bottom=2pt, width=\linewidth, boxrule=1pt]
    {\small {\fontfamily{zi4}\selectfont
    Goal: Wash sneakers\\
    Context: I remove shoelaces. I rinse.\\
    Question: What is the likelihood that my feet get wet by wearing the sneakers?\\
    Answer: likely
    }
    \par}
    \end{tcolorbox}
    \end{center}
    \caption{Our GPT-3 prompt, which is typical for a QA task. Each likelihood label is compared with the previous one to get the label for the change.}
    \label{fig:GPT-3_prompt}
    \vspace{-.4cm}
\end{figure}

Table~\ref{table:baseline} shows that all state-of-the-art LLMs we have attempted achieve close-to-chance performance on CREPE around 0.350 F1, whereas \texttt{text-davinci-003} and \texttt{ChatGPT} which are known to be stronger at reasoning perform better. Details about prompt formulation and experimental results on prompt sensitivity are shown in Appendix~\ref{app:pe} and~\ref{app:GPT-3-exp}.

\begin{table*}[!t]
\centering
    \small
\begin{tabular}{l|ll|llllllll|l}
\toprule
   & \multicolumn{2}{c|}{\textbf{Naive}} & \multicolumn{8}{c|}{\textbf{Large Language Models}}    & \textbf{Human}  \\ 
   & Cha.      & Maj.     & T5 & T0 & GPT3C & GPT3C+S &GPT3D2 &GPT3D3 &ChatGPT & \makecell{Codex\\(ours)} &  \\
Params & -          & -           & 3B & 11B & 13B & 13B   &175B &175B  & 175B  & 175B & -            \\ \midrule
Dev  & .262         & .297        & .343 & .336 & .346 & .341 & .350& .424 & .470 & \textbf{.585} & .868 \\ 
Test & .251         & .296       & .343 & .337 & .356 & .346& .533  & .423 & .462  & \textbf{.591} & -  \\ \bottomrule
\end{tabular}
\caption{Macro F1 of baseline models on the CREPE dataset. Human performance is not benchmarked on the test set as we strictly hold out its labels during all experiments. GPT3C represents the \texttt{text-curie-001} model. GPT3D2 represents the \texttt{text-davinci-002} model with an abnormal performance on the test set that we have confirmed but regrettably cannot explain. GPT3D3 represents the \texttt{text-davinci-003} model. GPT3C+S represents the GPT-3 \texttt{curie} model finetuned on StrategyQA. All of the above models work with textual prompts. Codex represents the \texttt{code-davinci-002} model and works with our proposed code-like prompts.}
\label{table:baseline}
\vspace{-.3cm}
\end{table*}

\subsection{Representing Procedures as Python Code}
\label{section:prompt}
\textbf{Codex} \cite{chen2021evaluating} is a variation of GPT-3 that was designed to be prompted with and to generate code, in addition to natural language texts. Recently, \citet{madaan22emnlp} found that prompting Codex with some structured representation such as Python code. Inspired by this observation, we propose novel code representations of procedures and hypothetical events. Among many possibilities we experimented with, the representation with the best empirical performance is described below, later shown to greatly outperform all baseline models. The representation is exemplified in Figure~\ref{fig:codex_prompt}.

The procedure is represented as a class where the goal $G$ is the class name, followed by the steps $s_i$ as comments. Then, each step is defined as a member function, in which the hypothetical events $e_j$ are represented as objects with comments. Each event object has an attribute ``change'' whose value describes the change of the likelihood. During inference, Codex is provided with the prompt including three in-context examples and the current procedure up to the definition of the ``init'' function and predicts the definition of all step functions. Finally, we extract the assigned value of the ``change'' attribute as the event likelihood change $\delta_i$.

This prompt design effectively leverages the semantic similarity between procedures with entity states and functions with variables, by representing texts as function identifiers and comments. We use \texttt{code-davinci-002}\footnote{While OpenAI announced that \texttt{text-davinci-002} is based on \texttt{code-davinci-002} (\url{https://platform.openai.com/docs/model-index-for-researchers}), we empirically find the former to perform worse with our code prompt and thus only consider the latter with code prompt.} with 175B parameters and default hyperparameters with a temperature of 0.

\begin{figure}[!t]
    \centering
    \begin{tcolorbox} [top=2pt,bottom=2pt, width=\linewidth, boxrule=1pt]
\begin{tabular}{rp{15cm}}
\begin{lstlisting}[numbers=none, basicstyle=\fontfamily{cmtt}\small,style=pythoncode,belowskip=-\baselineskip,aboveskip=- 0.5\baselineskip,commentstyle=\color{codegreen}]
class Wash_Sneakers:
  # Init
  # Remove shoelaces
  # Rinse
  def __init__(self, event0):
    self.event0 = event0 # My feet get wet by wearing the sneakers.
  def remove_shoelaces(self):
    self.event0.change = "equally likely" # My feet get wet by wearing the sneakers.
  def rinse(self):
    self.event0.change = "more likely" # My feet get wet by wearing the sneakers.
\end{lstlisting}
\end{tabular}
\end{tcolorbox}
    \caption{Our best-performing Python code representation of a procedure and hypothetical events, for Codex.}
    \label{fig:codex_prompt}
\vspace{-.3cm}
\end{figure}

\subsection{Results}
As CREPE is a ternary classification task, we report the macro F1 score across the three classes. As shown in Table~\ref{table:baseline}, T5 and T0 perform only slightly better (.343 and .336 F1) than chance (.297 F1). GPT-3, one of the most dominant models across a variety of NLP tasks, is no better (.336 F1), whereas finetuning it on another multihop reasoning dataset StrategyQA does not bring about any improvement (.341 F1). The latest GPT-3 models, \texttt{text-davinci-003} (.424 F1) and \texttt{ChatGPT} (.470 F1) which were released contemporarily with this paper, greatly outperform their predecessors. 

On the other hand, our code-representation of events as the prompt to Codex greatly outperforms all other models with .585 F1. As Codex is trained on public Github code in addition to the internet texts that GPT-3 is trained on, it is noteworthy that Codex can effectively reason about texts with code-like structures, for a procedure has many analogies to a class in object-oriented programming. 

\begin{table}[t!]
\centering
\small
\begin{tabular}{lll}
\toprule
                  & \textbf{Dev} & \textbf{Test} \\ \midrule
Codex             &  \textbf{.585}   &  \textbf{.591}    \\
no step comments    & .377    &   .352   \\
no event comments   &  .576  &   .555   \\
nested function   &  .568   &   .572   \\
flat variables &   .338  &   .341   \\ \bottomrule
\end{tabular}
\caption{Macro F1 of the ablations of our Codex prompt.}
\label{table:ablation}
\vspace{-.3cm}
\end{table}

\subsection{Ablation Studies}
To understand why the representation in our Codex prompt is effective, we perform an ablation study with various changes of the format to the representation, including:
\begin{itemize}[leftmargin=*]
    \item Remove steps comments in the beginning
    \item Remove event comments in step functions
    \item Use nested functions instead of a class
    \item Use flat variables to encode goals, steps, and events (no hierarchical class functions)
\end{itemize}
Examples of these empirically inferior representations are shown in Appendix~\ref{app:pe}. As seen in Table~\ref{table:ablation}, the hierarchical representation of procedures, steps, and events as classes or nested functions is critical. Besides, listing all the steps as comments helps, mimicking a programmer's textual explanation of a class or a function. 

\section{Causal Reasoning with Entities}

When a human tries to predict whether the event ``\textit{one would get burnt by touching a pan}'' is likely, their reasoning process would first focus on some entities in the question~(e.g., ``the pan''), then attend to some attributes and states of that entity~(e.g., the temperature of the pan is hot), and finally draw a logical conclusion~(e.g., ``the pan being hot means one would get burnt by touching it.'') CREPE is constructed precisely with this thought process in mind. An entity-attribute-change tuple is annotated along with each event likelihood change. In this section, we study how to explicitly leverage the intermediate information to assist the prediction of event likelihood prediction.

\subsection{Predicted Entity States as CoT} \label{sec:cot}
In CREPE, the task of predicting event likelihood change can be seen as a case of multihop reasoning, where a model first decomposes the question into some open-ended sub-questions, answer these sub-questions, and aggregate them as a final answer. LLMs can be prompted to perform chain-of-thought (CoT) style reasoning \cite{nye2021show,wei2022chain}. Thus, we ask the question:
\begin{displayquote}
\textbf{Q1}. Can LLMs benefit from first \textbf{predicting} entity state changes, as a CoT, before predicting event likelihood changes?
\end{displayquote}

\paragraph{CoT with GPT-3.}
First, we prompt GPT-3 with \citet{wei2022chain}'s CoT paradigm and \citet{pressmeasuring}'s self-ask paradigm, both of which are shown in Figure~\ref{fig:GPT-3_prompt_mh}. While self-ask relies on search engines for fact retrieval, we use LM generation instead as most of our entity state tracking questions are heavily context-dependent and unanswerable by any search engine. When writing demonstrations for few-shot learning, we impose the following logic progression for the follow-up questions: (1) initial followups shall ask questions on the state of entities that are directly related to the event; (2) followups following the entity state questions shall ask for the logical relationship between the entity states and the original event.

\begin{figure}[!t]
    \centering
    \begin{center}
    
    \begin{tcolorbox} [top=2pt,bottom=2pt, width=\linewidth, boxrule=1pt]
    {\small {\fontfamily{zi4}\selectfont
    Goal: Wash sneakers\\
    Context: I remove shoelaces. I rinse.\\
    Question: What is the likelihood that my feet get wet by wearing the sneakers?\\
    Answer: To get feet wet by wearing the sneakers, the sneakers must be wet. In the given context, the sneakers are wet. Therefore, comparing to the previous step, the likelihood change is "more likely".
    }
    \par}
    \end{tcolorbox}
    
    \begin{tcolorbox} [top=2pt,bottom=2pt, width=\linewidth, boxrule=1pt]
    {\small {\fontfamily{zi4}\selectfont
    Goal: Wash sneakers\\
    Context: I remove shoelaces. I rinse.\\
    Question: What is the likelihood that my feet get wet by wearing the sneakers?\\
    Follow up: Are the sneakers wet?\\
    Intermediate answer: Yes\\
    Follow up: Will my feet get wet by wearing wet sneakers?\\
    Intermediate answer: Yes \\
    Answer: likely
    }
    \par}
    \end{tcolorbox}    
    \end{center}
    \caption{Our GPT-3 prompt with intermediate questions, mimicking the CoT prompt (top) and the Self-Ask prompt (bottom).}
    \label{fig:GPT-3_prompt_mh}
    \vspace{-.3cm}
\end{figure}

\begin{table*}[]
\centering
    \small
\begin{tabular}{c|c|cc|cccc|c}
\toprule
   & \multicolumn{1}{c|}{\textbf{Naive}} & \multicolumn{2}{c|}{\textbf{LLMs}} & \multicolumn{4}{c|}{\textbf{CoT Large Language Models}}    & \textbf{Human}  \\ \midrule
   & Majority     & GPT-3   & Codex     & GPT-3 + CoT & GPT-3+self-ask & \makecell{Codex soft\\(ours)} & \makecell{Codex hard\\(ours)}  &  \\
Dev & .297     & .346     & .585        &  0.359 & .342 & .624   & \textbf{.667  }  & .868   \\ 
Test & .296    & .356      & .591       & 0.379    & .345 & \textbf{.626}   & .609   & -  \\\bottomrule
\end{tabular}
\caption{Macro F1 of chain-of-thought models on the CREPE dataset. GPT-3 + CoT|self-ask represents the \texttt{text-davinci-002} model prompted with the CoT or self-ask style prompt. }
\label{table:baseline_cot}
\vspace{-.3cm}
\end{table*}

\paragraph{CoT Codex with Soft Entity Representation.}
We modify our Codex prompt in Figure~\ref{fig:codex_prompt}, so that a sub-event is represented as a string variable whose declaration and value assignments are right before those of the hypothetical event. We refer to this as a \textit{soft representation} of entities (Figure~\ref{fig:codex_prompt_w_event}). 
During inference, Codex is provided with the code up to the step function header and predicts the entity and event changes for every step function. Our Codex model achieves the new best performance of .624 F1, outperforming the same model without predicted entities as CoT by .039 F1.

\begin{center}
\begin{tcolorbox} [top=2pt,bottom=2pt, width=\linewidth, boxrule=1pt]
\begin{tabular}{rp{15cm}}
\begin{lstlisting}[numbers=none, basicstyle=\fontfamily{cmtt}\small,style=pythoncode,belowskip=-\baselineskip,aboveskip=- 0.5\baselineskip,commentstyle=\color{codegreen}]
class Wash_Sneakers():
  # Init
  # Remove shoelaces
  # Rinse
  def init(self, event0, subevent0):
    self.event0 = event0 # My feet get wet by wearing the sneakers.
    self.event0.subevent = subevent0 # The sneakers are wet
  def remove_shoelaces(self):
    self.event0.subevent.change = 
      "equally likely" # The sneakers are wet
    self.event0.change = "equally likely" # My feet get wet by wearing the sneakers.
  def rinse(self):
    self.event0.subevent.change = 
       "more likely" # The sneakers are wet
    self.event0.change = "more likely" # My feet get wet by wearing the sneakers.
\end{lstlisting}
\end{tabular}
\end{tcolorbox}
\end{center}
\vspace{-5mm}
\begin{figure}[h!]
    \centering
    \caption{Our Codex prompt with a soft representation of entity state changes as strings.}
    \label{fig:codex_prompt_w_event}
    \vspace{-.3cm}
\end{figure}

\paragraph{CoT Codex with Hard Entity Representation.}
The two approaches above both \textit{softly} represent the intermediate entity state changes as texts, either questions or statements. Here, LLMs are not enforced to generate intermediate reasoning steps that contain entities and attributes. To answer Q1 more precisely, we experiment with a \textit{hard entity representation} where the entity-attribute-change tuple is explicitly baked into the Codex prompt as shown in Figure~\ref{fig:codex_prompt_w_entity}. Here, each entity is represented as an object with an attribute and assigned value. The hard entity representation leads to a far superior performance of .667 F1 on the development set but generalizes worse on the test set with .609 F1. 

\begin{figure}[h!]
    \centering
    \begin{tcolorbox} [top=2pt,bottom=2pt, width=\linewidth, boxrule=1pt]
\begin{tabular}{rp{15cm}}
\begin{lstlisting}[numbers=none, basicstyle=\fontfamily{cmtt}\small,style=pythoncode,belowskip=-\baselineskip,aboveskip=- 0.5\baselineskip,commentstyle=\color{codegreen}]
class Wash_Sneakers():
  # Init
  # Remove shoelaces
  # Rinse
  def init(self, event0):
    self.sneakers = Sneakers()
    self.event0 = event0 # My feet get wet by wearing the sneakers.
  def remove_shoelaces(self):
    self.event0.change = "equally likely" # My feet get wet by wearing the sneakers.
  def rinse(self):
    self.sneakers.wet = True
    self.event0.change = "more likely" # My feet get wet by wearing the sneakers.
\end{lstlisting}
\end{tabular}
\end{tcolorbox}
    \caption{Our Codex prompt with a hard representation of entity states as variables, attributes, and values.}
    \label{fig:codex_prompt_w_entity}
\end{figure}

To recap, we have shown that LLMs can be prompted to exhibit a CoT that first predicts entity state changes and then event likelihood changes. Hence, our answer to \textbf{Q1} raised at the beginning of this subsection is `yes.'

\begin{table}[t!]
\centering
    \small
\begin{tabular}{lll}
\toprule
                  & \textbf{Dev} & \textbf{Test} \\ \midrule
Majority             &  .297   &  .296    \\ \midrule
GPT-3 CoT             &  .342   &  .345    \\
w/ gold entity changes    & .351    &   .380   \\
Codex CoT   &  .667  &   .609   \\
w/ gold entity changes   &  \textbf{.715}   &   \textbf{.722}   \\ \midrule 
Human             &  .868   &  -    \\\bottomrule
\end{tabular}
\caption{Macro F1 of GPT-3 and Codex with chain-of-thought provided with gold entity state changes.}
\label{table:gold_entities}
\vspace{-.3cm}
\end{table}

\subsection{Annotated Entity States as CoT}
In the above section, we have shown how event likelihood prediction can be improved by first having the LLMs predict entity states as a CoT. These experiments mimic a realistic setting where information about entities is unavailable. However, in some scenarios, the entity states may be provided. For example, an embodied agent or a robot might have a reliable component that tracks entities; some practitioners might care about a small set of procedures in a narrow domain with annotated entity changes; or, some event schemata containing entity information could be used to predict unseen events. Here, we try to answer the following question:
\begin{displayquote}
\textbf{Q2}. Can LLMs effectively leverage \textbf{annotated} entity state changes to better predict event likelihood changes?
\end{displayquote}

Instead of having LLMs predict entity state changes, we provide the annotated entity state changes in the CREPE dataset to GPT-3 and Codex. Doing so has the additional benefit of verifying that entity state changes indeed causally benefit LLMs in predicting events.

As shown in Table~\ref{table:gold_entities}, our Codex representation with access to gold entity changes leads to improved performance of .715 F1 on the development set. In contrast, GPT-3 does not see any gain. Hence, the answer to \textbf{Q2} is `yes' for the code-trained LLMs but `no' for standard LLMs.

\subsection{Externally Predicted Entity States}

As we will discuss further in Section~\ref{sec:related_work}, entity state tracking is an established task in NLP with existing datasets and models. We have now predicted entity state changes using LLMs in a few-shot learning setting. It is then natural to pose the question:
\begin{displayquote}
\textbf{Q3}. Do existing entity state tracking models make predictions that lead to better performance on CREPE?
\end{displayquote}
Our definition of causal reasoning of events is directional since   we consider entity state changes as the cause of the change in event likelihoods. To this extent, we incorporate OpenPI \cite{tandon-etal-2020-dataset}, the only open-domain entity state tracking dataset in procedural texts, as a part of the pipeline. In OpenPI, the input is a goal, a step, and the output is tuples of an entity, a feature, and two attributes before and after the execution of the step. For example, after ``heat the pan [step]'', ``the temperature [feature] of the pan [entity] is cool [attribute] before and hot [attribute] afterward.'' While the original paper proposed a GPT2 model \cite{radford2019language}, we opt to finetune the superior GPT-3 Curie model on its data. After the model makes a prediction, we post-process it into the format of CREPE by discarding the feature and producing two entity-attribute-change pairs (e.g., pan-hot-``more likely'' and pan-cold-``less likely''). We provide Codex with only the entity changes when the entity is mentioned in the event. Further, to fit our prompt in the context window of Codex, we provide Codex with 5 entity state changes uniformly drawn from a pool of candidate choices at every step. The resulting OpenPI-prompted Codex gives a degraded macro F1 score of 0.553 on the development set and 0.496 on the testing set. Hence, our answer to \textbf{Q3} is `no,' suggesting that existing entity state tracking datasets may be insufficient for our causal reasoning task.

\section{Performance Analysis}
In this section, we analyze potential factors that play a role in our Codex model's performance. We investigate three factors: (1) the number of steps in a procedure; (2) explicit mentions of event-related entity-of-interest (EoI) in a given step; and (3) the logical relation (entailment or contradiction) between the event likelihood change and its related entity state change. To study factor (1), we dichotomize procedures from the development set by the average length of the procedure. To investigate factors (2) and (3), we manually labeled the ground truth EoI mentioning and logical relation for the development dataset. Intuitively, estimating event likelihood in lengthy procedures and in steps where EoI is not explicitly mentioned would be difficult. Rather surprisingly, Codex shows no significant performance discrepancy under factors (2) and (3), and only a slight performance difference in factor (1) (see Appendix~\ref{app:ea}). %These results suggest that Codex is adept at comprehending complicated temporal information and recalling previously mentioned content. 

Further, the task of CREPE can be divided into two sub-tasks, first to identify whether an event likelihood change occurred at all, and then to classify the change as either more or less likely. 
%Figure~\ref{fig:codex_cot} shows a comparison of macro F1 score on the two sub-tasks between Codex with and without CoT.
We observe that CoT Codex outperforms Codex on both sub-tasks. For the classification task, in particular, CoT Codex obtained a .149 increase in macro F1 score from .805 to .954. This shows not only that CoT Codex is effective, but also that its bottleneck is identifying event likelihood change.

\iffalse
\begin{figure}
    \centering
    \includegraphics[scale = 0.25]{images/error_analysis.png}
    \caption{Macro F1 score of identification and classification tasks of Codex with and without CoT prompting.}
    \label{fig:codex_cot}
\end{figure}
\fi

\section{Related Work}
\label{sec:related_work}

\paragraph{Event \& Entity Extraction and Representation}
Event-centric NLP has been a dominant strand of approaches to machine reasoning. Myriad work has focused on extracting events from the news or web data \cite{liu-etal-2018-jointly,yang-etal-2019-exploring,du-cardie-2020-event}. The effort of structurally representing scripts, groups of events in certain scenarios including procedures, started decades ago \cite{abelson1977scripts} and is receiving revived attention in present years \cite{li-etal-2020-connecting,wang2022schema}. While this line of work mostly focuses on the representation as relations (e.g., temporal, hierarchical) among events, we recognize entities as a cause of event relations and thus propose a more granular representation. Furthermore, structured representations of events typically cannot take advantage of the power of textual LLMs for challenging downstream tasks. In contrast, we advance towards the best of two worlds by working with code language models.

Besides, existing work on jointly extracting and representing events and entities \cite{lee-etal-2012-joint,wadden-etal-2019-entity,barhom-etal-2019-revisiting} neglects the causal relation therein and treats entities and events simply as two related tasks to be tackled simultaneously. We causally bridge the two.

\paragraph{Entity State Tracking}

Prior work on entity state tracking spans various disciplines of AI. For instance, object tracking, a sub-task of entity state tracking, has led to much work in both robotics \cite{wang2007simultaneous} and computer vision \cite{comaniciu2003kernel}. In NLP, early efforts focus on synthetic, closed-domain data \cite{weston2015towards, long-etal-2016-simpler} and more recent ones shift attention to real-world procedures \cite{bosselut2017simulating, dalvi-etal-2018-tracking,gupta-durrett-2019-tracking,du-etal-2019-consistent,mysore-etal-2019-materials} with a closed set of entities and attributes or an open-ended set \citet{tandon-etal-2020-dataset}. In all prior work, entity state track is treated as an end-task, whereas we treat it as a critical intermediate step for event reasoning, a more practical application. 

\paragraph{Counterfactual Reasoning}
In this work, we hope to provide evidence that signals of entities effectively help models reason about events. We specifically focus on hypothetical event reasoning because it is a high-level cognitive ability beyond pattern recognition and a manifestation of complex reasoning ability \cite{10.5555/3238230, pearl2019seven}. Counterfactual reasoning has a long history with formal methods \cite{forbus1984qualitative,lewis2013counterfactuals}. Less modern work exists in commonsense \cite{feng-etal-2021-empowering}, procedural texts \cite{tandon-etal-2019-wiqa}, and even computer vision \cite{yue2021counterfactual}. 

\paragraph{Multihop Reasoning}
Prior studies on multihop reasoning mainly focus on question answering from a passage \cite{welbl2018constructing, talmor-berant-2018-web, yang-etal-2018-hotpotqa, kovcisky2018narrativeqa, mihaylov2018can, khot2020qasc} and representing and utilizing multihop information in the form of structured data \cite{de-cao-etal-2019-question, ding-etal-2019-cognitive, qiu2019dynamically, cao-etal-2019-bag, fang-etal-2020-hierarchical, thayaparan-etal-2019-identifying, zhang2020coarse, zhang2021answering, huang2021breadth}.

There are also efforts such as DecompRC, StrategyQA, and CGDe-FGIn that attempt to conduct multihop reasoning by decomposing the original task to a series of logically related sub-tasks \cite{min-etal-2019-multi, geva-etal-2021-aristotle, cao2022coarse}. Such an approach has recently seen great success with the Chain-of-Thought (CoT) prompting of GPT-3, which significantly improves numerous multihop reasoning tasks \cite{nye2021show, kojima2022large, wei2022chain, wang2022self}. Following CoT prompting, Self-Ask further elicits CoT by demanding GPT-3 to explicitly generate the reasoning questions raised during its chain-of-thought process \cite{pressmeasuring}.

\paragraph{Code-Based Language Models and Prompts}
Recent work has shown that LLMs trained on programs or code (PLMs) have an augmented ability of reasoning over natural language texts. Notably, \citet{suzgun2022challenging,liang2022holistic} showed that PLMs outperforms only-text-trained LMs on certain reasoning tasks even though the prompts are purely natural language and contain no code. Moreover, there has been speculation that multihop reasoning is an emergent ability exclusive to PLMs and absent in their only-text-trained predecessors \cite{fu2022gptroadmap}. 

Even more interestingly, a line of contemporary work found that, for some reasoning tasks, prompting PLMs with certain structured programs (e.g., Python code, JSON, PDDL) that represent the originally textual data outperforms doing so simply with natural language prompts. These tasks include math questions \cite{chen2022program, lyu-etal-2023,mishra2022lila} and event reasoning \cite{madaan22emnlp,wang2022code4struct} like our work.

\paragraph{Procedural Texts}
Procedural texts are an attractive data source to reason about events and entities which undergo frequent changes. There has been steady efforts in computer vision \cite{miech2019howto100m}, robotics \cite{ahn2022can}, and language \cite{9070972,https://doi.org/10.48550/arxiv.2205.07455}. In NLP specifically, work on procedures includes extracting them from instructional texts \cite{10.1145/584955.584977,delpech-saint-dizier-2008-investigating, zhang-etal-2012-automatically}, reasoning about events \cite{takechi-etal-2003-feature,tandon-etal-2019-wiqa,rajagopal-etal-2020-ask,zhang-etal-2020-reasoning}, knowledge-base construction \cite{jung2010automatic, chu2017distilling,park2018learning}, or applying them to downstream applications \cite{yang-etal-2021-visual, yang2021induce,zhang-etal-2020-analogous,lyu-etal-2021-goal,dalvi-etal-2019-everything,zhang-etal-2020-intent, chen-etal-2020-hybridqa}. Our work is scoped in procedural texts due to the outstanding causal relations between entities and events in a dynamic environment.

\section{Conclusion and Future Work}
We present CREPE, a benchmark for causal reasoning about events and entities in procedural texts. We show that mainstream LLMs such as GPT-3 perform close to chance on CREPE, while using code-like event representation as a prompt to code language model Codex greatly improves the performance. Further, we experiment with various ways to encode entity information into this representation and find that eliciting chain-of-thought reasoning from Codex further improves performance while existing CoT approaches with GPT-3 are ineffective. We clearly show that LLMs benefit from lower-level entity information when making predictions about higher-level events. Future work should explore related tasks such as next-event prediction, event temporal ordering, etc., by injecting relevant information about entities into our representation. Our code-representation of events allows more powerful expressions than simply entailment and negation considered in this work. Future work may explore other forms of code chain-of-thought such as first-order logic. These expressions generated by LLMs can be computed objectively, thus ameliorating LLMs' hallucinations and improving the interpretability and faithfulness of predictions.

\section{Limitations}
Despite our best efforts, our CREPE dataset has inherent limitations. First, the choice of studying procedure texts, despite many discussed advantages, limits the domain, writing style, and other semantic features of the texts. As a result, porting our methods and findings to other text styles such as stories or news might require domain adaptation. Second, we prioritize quality over quantity when creating this benchmark, which suffers from small size and contains biases from the annotators, even though we address the latter by having different annotators label a test set. 

When annotating the hypothetical events, our intention is that they represent a wild variety that doers of the procedures, humans or machines, would care about. However, we also have to ensure these events are unambiguously bound to some entities in order to challenge models for their causal reasoning ability. While we do our utmost to balance these two conflicting objectives, the issue might still persist.

In CREPE, each event likelihood change is caused by exactly one entity state change. This is an over-simplification made to facilitate evaluation. In real life, many complex events require many entity states to be reasoned about, which in turn may have complex logical relations among them. We leave this for future work.

While we intend our representation of events and entities to be a general and effective one, we have only shown that it works well empirically using Codex, which is one of the only code language models at present. Whether the idea of our structured representation applies to other models remains to be explored.

\section{Acknowledgements}
This research is based upon work supported in part by the DARPA KAIROS Program (contract FA8750-19-2-1004), the DARPA LwLL Program (contract FA8750-19-2-0201), the IARPA BETTER Program (contract 2019-19051600004), and the NSF (Award 1928631). Approved for Public Release, Distribution Unlimited. The views and conclusions contained herein are those of the authors and should not be interpreted as necessarily representing the official policies, either expressed or implied, of DARPA, IARPA, NSF, or the U.S. Government. 

We thank the students in the Artificial Intelligence class at the University of Pennsylvania in Fall 2022 who participated in the annotation of the test set of CREPE. We thank Niket Tandon and Qing Lyu for valuable discussions about this work.

\bibliography{anthology,custom}
\bibliographystyle{acl_natbib}
\clearpage
\appendix
\section{Prompt Sensitivity}
\label{app:GPT-3-exp}
In addition to the results reported in Table~\ref{table:baseline}, we also investigated the effect of the number and choice of in-context examples.

\paragraph{Number of in-context examples} 
The context window of \texttt{text-davinci-002} maximally fits 3 shots. We experiment with 1-shot (0.245 f1), 2-shots (0.348 f1), and 3-shots (0.359 f1) learning using text-002 with CoT prompting. We see that having more context provides limited improvements in model performance.

\paragraph{Prompt sensitivity with random examples}
We tested the \texttt{text-davinci-002} model with CoT prompt on the dev set using randomly chosen examples from our example bank. The F1 scores for 5 runs with randomly chosen in-context examples are 0.333, 0.327, 0.359, 0.336, and 0.331. The mean score is 0.337, and the standard deviation is 0.011, implying low sensitivity of in-context example selection.

\section{Prompt Engineering}
\label{app:pe}
\subsection{Code Prompts for Codex}
In Section 4 and 5, we have discussed our best-performing prompts for GPT-3 and Codex. Here, we elaborate on inferior Codex prompts and shed light on why they do not work well empirically.

\paragraph{Best prompt} 
%1.2
As discussed, our best-performing prompt represents procedures as classes and steps as functions.

\begin{center}
\begin{tcolorbox} [top=2pt,bottom=2pt, width=\linewidth, boxrule=1pt]
{\small {\fontfamily{zi4}\selectfont
\begin{verbatim}
class Wash_Sneakers:
  # Init
  # Remove shoelaces
  # Rinse
  def __init__(self, event0):
    self.event0 = event0 # My feet get wet by
    wearing the sneakers.
  def remove_shoelaces(self):
    self.event0.change = "equally likely" # My
    feet get wet by wearing the sneakers.
  def rinse(self):
    self.event0.change = "more likely" # My
    feet get wet by wearing the sneakers.
\end{verbatim}
}
\par}
\end{tcolorbox}
\end{center}

\paragraph{Nested functions} 
%1.2.4
Instead of representing procedures as classes as in our best-performing prompt, we can also represent them as nested functions. 

\begin{center}
\begin{tcolorbox} [top=2pt,bottom=2pt, width=\linewidth, boxrule=1pt]
{\small {\fontfamily{zi4}\selectfont
\begin{verbatim}
def wash_sneakers(event0):
  # Init
  # Remove shoelaces
  # Rinse
  event0 = event0 # My feet get wet by
    wearing the sneakers.
  def remove_shoelaces(self):
    event0.change = "equally likely" # My
    feet get wet by wearing the sneakers.
  def rinse(self):
    event0.change = "more likely" # My
    feet get wet by wearing the sneakers.
\end{verbatim}
}
\par}
\end{tcolorbox}
\end{center}

\paragraph{No step comments} 
%1.2.1
The comments displaying the steps immediately after the class declaration are removed.

\begin{center}
\begin{tcolorbox} [top=2pt,bottom=2pt, width=\linewidth, boxrule=1pt]
{\small {\fontfamily{zi4}\selectfont
\begin{verbatim}
class Wash_Sneakers:
  def __init__(self, event0):
    self.event0 = event0 # My feet get wet by
    wearing the sneakers.
  def remove_shoelaces(self):
    self.event0.change = "equally likely" # My
    feet get wet by wearing the sneakers.
  def rinse(self):
    self.event0.change = "more likely" # My
    feet get wet by wearing the sneakers.
\end{verbatim}
}
\par}
\end{tcolorbox}
\end{center}

\paragraph{No event comments} 
%1.2.2
The comments displaying the events in step functions except init are removed.

\begin{center}
\begin{tcolorbox} [top=2pt,bottom=2pt, width=\linewidth, boxrule=1pt]
{\small {\fontfamily{zi4}\selectfont
\begin{verbatim}
class Wash_Sneakers:
  def __init__(self, event0):
    self.event0 = event0 # My feet get wet by
    wearing the sneakers.
  def remove_shoelaces(self):
    self.event0.change = "equally likely"
  def rinse(self):
    self.event0.change = "more likely"
\end{verbatim}
}
\par}
\end{tcolorbox}
\end{center}

\paragraph{Two-step} In this approach, we hypothesize that providing entity state change at every step is helpful. To do this, we first prompt Codex to generate entity states corresponding to a specific event:

\begin{center}
\begin{tcolorbox} [top=2pt,bottom=2pt, width=\linewidth, boxrule=1pt]
{\small {\fontfamily{zi4}\selectfont
\begin{verbatim}
class Wash_Sneakers:
    def remove_shoelaces(self):
        event = "My feet get wet by wearing 
                the sneakers."
        event.precondition = \ 
                    ("sneakers", "wet")
    def rinse(self):
        event = "My feet get wet by wearing 
                the sneakers."
        event.precondition = \
                    ("sneakers", "wet")
\end{verbatim}
}
\par}
\end{tcolorbox}
\end{center}

We select event-related entities by majority vote. The resulting entity state bank is used to prompt Codex to first deduce entity state at every step and then answer the likelihood of the event.

\paragraph{Flat variables} Instead of defining functions using \texttt{def} or creating class with \texttt{class}, we use only variables to define relevant information.
\begin{center}
\begin{tcolorbox} [top=2pt,bottom=2pt, width=\linewidth, boxrule=1pt]
{\small {\fontfamily{zi4}\selectfont
\begin{verbatim}
Goal = "Wash Sneakers"

Context = "Remove shoelaces. After this, 
           the shoelaces are removed"
Question = "What is the likelihood that my feet
            get wet by wearing the sneakers?
Options = [
           "more likely", 
           "less likely", 
           "equally likely",
          ]
Answer = Options[2]

Context = "Rinse the sneakers. After this, 
          the sneakers are damp."
Question = "What is the likelihood that my feet
           get wet by wearing the sneakers?
Options = [
           "more likely", 
           "less likely", 
           "equally likely",
          ]
Answer = Options[0]
\end{verbatim}
}
\par}
\end{tcolorbox}
\end{center}

\subsection{Textual Prompts for GPT-3}
For GPT-3, we attempted a dozen of prompt formulations in our preliminary experiments which we found to differ minimally in performance. Here, we show one example:
\begin{center}
\begin{tcolorbox} [top=2pt,bottom=2pt, width=\linewidth, boxrule=1pt]
{\small {\fontfamily{zi4}\selectfont
\begin{verbatim}
"Wash hands" involves the followings steps:
1. Turn on the tap water.
2. Put hands under running water.
3. Apply soap and rub hands.
4. Turn off the tap water.
5. Dry my hands using a towel.

For every step, find out how likely it is that
water streaming sound can be heard. Answer as
(A) very likely (B) likely (C) not very likely
(D) unlikely.  

Step 1: (A) very likely
Step 2: (A) very likely
Step 3: (A) very likely
Step 4: (D) unlikely
Step 5: (D) unlikely
\end{verbatim}
}
\par}
\end{tcolorbox}
\end{center}

For GPT-3 finetuned with StrategyQA, we ask two questions regarding the likelihood of the events, namely whether it is more/less likely that some event occurs. After obtaining the result, we conduct a consistency check. For consistent likelihood estimates, where only one of the two questions gives a positive answer, or both questions give negative answers, we assign the corresponding label to the event state change. For inconsistent estimates, where both questions give positive answers, we assign the event change likelihood to the majority label, which is "equally likely". An example of a finetuning prompt-completion pair is shown as follows

\begin{center}
\begin{tcolorbox} [top=2pt,bottom=2pt, width=\linewidth, boxrule=1pt]
{\small {\fontfamily{zi4}\selectfont
\begin{verbatim}
Prompt:
Context: Julius Caesar had three children. 
Genghis Khan had sixteen children. 
Modern geneticists have determined 
that out of every 200 men today 
has DNA that can be traced to 
Genghis Khan.
Question: Are more people today 
related to Genghis Khan than Julius Caesar?
Take it step by step:

Completion:
#1 How many kids did Julius Caesar have?
two
#2 How many kids did Genghis Khan have?
fourth
#3 Is fourth greater than two?
no
Therefore, the answer to the original 
question is True
\end{verbatim}
}
\par}
\end{tcolorbox}
\end{center}

An example of our StrategyQA GPT-3 prompt on the CREPE task is as follows:

\begin{center}
\begin{tcolorbox} [top=2pt,bottom=2pt, width=\linewidth, boxrule=1pt]
{\small {\fontfamily{zi4}\selectfont
\begin{verbatim}
Context: Remove shoelaces. Rinse. Srub the
shoes with cleaning solution. Rinse the shoes 
again. Air dry the shoes and put the shoelaces 
back on.
Question: Is it more likely that my feet get
wet by wearing the sneakers?
Take it step by step:

Completion:
#1 Is the sneaker wet?
Yes
#2 Will my feet get wet by wearing wet shoes?
Yes
Therefore, the answer to the original question
is True.
\end{verbatim}
}
\par}
\end{tcolorbox}
\end{center}

\subsection{Textual Prompts for ChatGPT}
As of the time of camera-ready submission of this paper (Feburary 1, 2023), OpenAI has not released the API for ChatGPT. Thus, we use an unofficial API\footnote{\url{https://github.com/acheong08/ChatGPT}} which is believed to behave the same as the official web playground. Because ChatGPT is designed to only work with a zero-shot and multi-turn dialog setting, we tweak our prompt as follows:

\begin{center}
\begin{tcolorbox} [top=2pt,bottom=2pt, width=\linewidth, boxrule=1pt]
{\small {\fontfamily{zi4}\selectfont
\begin{verbatim}
I'm trying to wash hands.
First, I turn on the tap water.
At this point, is it likely that
water streaming sound can be heard?
Answer with yes or no.
[answer]
Then, I put hands under running water.
At this point, is it likely that
water streaming sound can be heard?
Answer with yes or no.
[answer]
...
\end{verbatim}
}
\par}
\end{tcolorbox}
\end{center}

\subsection{Textual Prompts for T5/T0}
We design the following prompt for T5 and T0 to perform our task:
\begin{center}
\begin{tcolorbox} [top=2pt,bottom=2pt, width=\linewidth, boxrule=1pt]
{\small {\fontfamily{zi4}\selectfont
\begin{verbatim}
Goal: [The name of the goal]
Step: [The list of steps]
Question: Is that okay that [question]?
Answer: [yes or no, generated by the model]
\end{verbatim}
}
\par}
\end{tcolorbox}
\end{center}

\section{Error Analysis}
\label{app:ea}
In Section 6, we conclude that the performance of Codex is not influenced by (1) the number of steps in a procedure; (2) explicit mentions of event-related entity-of-interest (EoI) in a given step; and (3) the logical relation (entailment or contradiction) between the event likelihood change and its related entity state change. 

\begin{table}[h!]
\centering
\begin{tabular}{ll}
\toprule
Factors   & Dev \\ 
\midrule
Procedure Length > 7    &   .629   \\
Procedure Length $\leq$ 7    &  .700   \\
\midrule
EoI Mentioned & .481 \\
EoI NOT Mentioned & .496\\
\midrule
Entailment   &  .482   \\
Contradiction &  .461   \\ 
\bottomrule
\end{tabular}
\caption{Macro F1 Score of error analysis. The scores for EoI and Logical relation are lower since we do not consider the majority label, "equally likely", in the error analysis.}
\label{tab:bad_prompts}
\end{table}

\end{document}